\documentclass{llncs}

\usepackage{graphicx}

\begin{document}

\pagestyle{empty}

\mainmatter

\title{Traffic Flow Forecasting Using a Spatio-Temporal Bayesian
Network Predictor}

\titlerunning{Lecture Notes in Computer Science}

\author{
 Shiliang Sun, Changshui Zhang, and Yi Zhang}

\authorrunning{Shiliang Sun et al.}

\institute{State Key Laboratory of Intelligent Technology and
Systems,\\ Department of Automation, Tsinghua University, Beijing,
China, 100084\\
\email{sunsl02@mails.tsinghua.edu.cn,
\{zcs,zhyi\}@mail.tsinghua.edu.cn}}

\maketitle

\begin{abstract}
A novel predictor  for traffic flow forecasting, namely
spatio-temporal Bayesian network predictor, is proposed. Unlike
existing methods, our approach incorporates all the spatial and
temporal information available in a transportation network to
carry our traffic flow forecasting of the current site. The
Pearson correlation coefficient is adopted to rank the input
variables (traffic flows) for prediction, and the best-first
strategy is employed to select a subset as the cause nodes of a
Bayesian network. Given the derived cause nodes and the
corresponding effect node in the spatio-temporal Bayesian network,
a Gaussian Mixture Model is applied to describe the statistical
relationship between the input and output. Finally, traffic flow
forecasting is performed under the criterion of Minimum Mean
Square Error (M.M.S.E.). Experimental results with the urban
vehicular flow data of Beijing demonstrate the effectiveness of
our presented spatio-temporal Bayesian network predictor.
\end{abstract}

\section{Introduction}
Short-term traffic flow forecasting, which is to determine the
traffic volume in the next time interval usually in the range of
five minutes to half an hour, is an important issue for the
application of Intelligent Transportation Systems (ITS)
\cite{ex1}. Up to the present, some approaches ranging from simple
to elaborate on this theme were proposed including those based on
neural network approaches, time series models, Kalman filter
theory, simulation models, non-parametric regression, fuzzy-neural
approach, layered models, and Markov Chain models
\cite{ex1}$\sim$\cite{ygq03}. Although these methods  have
alleviated difficulties in traffic flow modelling and forecasting
to some extent, from a careful review we can still find a problem,
that is, most of them have not made good use of spatial
information from the viewpoint of networks to analyze the trends
of the object site. Though Chang et al utilized the data from
other roadways to make judgmental adjustments, the information was
still not used to its full potential \cite{chang}. Yin et al
developed a fuzzy-neural model to predict the traffic flows in an
urban street network whereas it only utilized the upstream flows
in the current time interval to forecast the selected downstream
flow in the next interval \cite{yinhb}.

The main contribution of this paper is that we proposed an
original spatio-temporal Bayesian network predictor, which
combines the available spatial information with temporal
information in a transportation network to implement traffic flow
modelling and forecasting. The motivation of our approach is very
intuitive. Although many sites may be located at different even
distant parts of a transportation network, there exist some common
sources influencing their own traffic flows. Some of the
distributed sources include shopping centers, home communities,
car parks, etc. People's activities around these sources usually
obey some consistent laws in a long time period, such as the
usually common working hours. To our opinion, these hidden sources
imply some information useful for traffic flow forecasting in
different sites. Therefore, construct a causal model (Bayesian
network) among different sites for traffic flow forecasting is
reasonable. This paper covers how to use the information from a
whole transportation network to design feasible spatio-temporal
Bayesian networks and carry our traffic flow forecasting of the
object sites. Encouraging experimental results with real-world
data show that our approach is rather effective for traffic flow
forecasting.

\section{Methodology} 

In a transportation network, there are usually a lot of sites
(road links) related or informative to the traffic flow of the
current site from the standpoint of causal Bayesian networks.
However, using all the related links as input variables (cause
nodes) would involve much irrelevance, redundancy and would be
prohibitive for computation. Consequently, a variable selection
procedure is of great demand. Up to date many variable selection
algorithms include variable ranking as a principal or auxiliary
selection mechanism because of its simplicity, scalability, and
good empirical success \cite{isabelle}. In this article, we also
adopt the variable ranking mechanism, and the Pearson correlation
coefficient is used as the specific ranking criterion defined for
individual variables.
\subsection{Variable Ranking and Cause Node Selection}

Variable ranking can be regarded as a filter method: it is a
preprocessing step, independent of the choice of the predictor
\cite{Kohavi}. Still, under certain independence or orthogonality
assumptions, it may be optimal with respect to a given predictor
\cite{isabelle}. Even when variable ranking is not optimal, it may
be preferable to other variable subset selection methods because
of its computational and statistical scalability \cite{hastie}.
This is also the motivation of our using the best-first search
strategy to select the most relevant traffic flows from the
ranking result  as final cause nodes of a Bayesian network.

Consider a set of $m$ samples $\{x_{k},y_{k}\} (k=1, ..., m)$
consisting of $n$ input variable $x_{k,i} (i=1, ..., n)$ and one
output variable $y_{k}$. Variable ranking makes use of a scoring
function $S(i)$ computed from the value $x_{k,i}$ and $y_{k} (k=1,
..., m)$. By convention, we assume that a high score is indicative
of a valuable variable and that we sort variables in decreasing
order of $S(i)$. Furthermore, let $X_{i}$ denote the random
variable corresponding to the $i^{th}$ component of input vector
$x$, and $Y$ denote the random variable of which the outcome $y$
is a realization. The Pearson correlation coefficient between
$X_{i}$ and $Y$ can be estimated by:
\begin{equation}
\label{PearsonEst}
R(i)=\frac{\sum_{k=1}^{m}(x_{k,i}-\overline{x}_{i})(y_{k}-\overline{y})}{\sqrt[]{\sum_{k=1}^{m}(x_{k,i}-\overline{x}_{i})^{2}\sum_{k=1}^{m}(y_{k}-\overline{y})^{2}}}
\end{equation}
where the bar notation stands for an average over the index $k$
\cite{isabelle}.

In this article, we use the norm $|R(i)|$ as a variable ranking
criterion. After the variable ranking stage, a variable selection
(cause node selection) process is adopted to determine the final
cause nodes (input variables) for predict the effect node
(output). Here we use the best-first search strategy to find the
cause nodes as the first several variables in the ranking list
because of its fastness, simplicity and empirical effectiveness.
\subsection{Flow Chart for Traffic Flow Forecasting}
Given the derived cause nodes and the effect node in a Bayesian
network, we utilize the Gaussian Mixture Model (GMM) and the
Competitive EM (CEM) algorithm to approximate their joint
probability distribution. Then we can obtain the optimum
prediction formulation as an analytic solution under the M.M.S.E.
criterion. For details about the GMM, CEM algorithm and the
prediction formulation, please refer to  articles
\cite{ygq03}\cite{ex9}\cite{isnn04}.

Now we describe the flow chart of our approach for traffic flow
forecasting. First the data set is divided into two parts, one
serving as training set for input variable (cause node) selection
and parameter learning, and the other test set. The flow chart can
be given as follows: 1) Choose an object road site whose traffic
flow should be forecasted (effect node) and collect all the
available traffic flows in a traffic network as the original input
variables; 2) Compute the Pearson correlation coefficients between
the object traffic flow (effect node) and the input variables on
the training set with different time lags respectively, and then
select several most related variables in the ranking list as the
final cause nodes of the spatio-temporal Bayesian network; 3)
Derive the optimum prediction formulation using GMM and CEM
algorithm detailed in articles \cite{ygq03}\cite{isnn04}; 4)
Implement  forecasting on the test set using the derived
formulation.

Conveniently, the flow chart can be largely reduced and for
real-time utility when forecasting a new traffic flow, because the
cause node selection  and the prediction formulation need only be
computed one time based on the historical traffic flows (learning
stage), and thus can be derived in advance.
\section{Experiments}
\begin{figure}[t]
\centerline{\includegraphics [height=5cm]{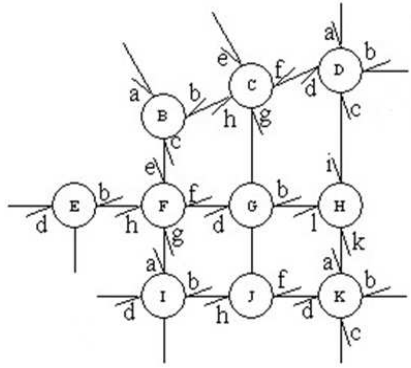}}
\caption{The analyzed transportation network } \label{usednet}
\end{figure}

The field data analyzed in this paper is the vehicle flow rates of
discrete time series recorded every 15 minutes along many road
links by the UTC/SCOOT system in Traffic Management Bureau of
Beijing, whose unit is vehicles per hour (veh/hr). Fig.
\ref{usednet} depicts a real patch used to verify our proposed
predictor. The raw data for utility are of 25 days and totally
2400 sample points taken from March, 2002. To validate our
approach, the frist 2112 points (training set) are employed to
carry out input cause node selection and to learn parameters of
the spatio-temporal Bayesian network, and the rest (test set) are
employed to test the forecasting performance.

In addition, we utilize the the Random Walk method and Markov
Chain method as base lines to evaluate our presented approach
\cite{ygq03}. Random Walk is a classical method for traffic flow
forecasting. Its core idea is to forecast the current value using
its last value, and can be formulated as:
\begin{equation}
\hat{x}(t+1)=x(t) \; .
\end{equation}
Markov Chain method models traffic flow as a high order Markov
chain. It has shown great merits over several other approaches for
traffic flow forecasting \cite{ygq03}. In this paper the joint
probability distribution for the Markov Chain method is also
approximated by the GMM whose parameters are estimated through CEM
algorithm. The number of input variables  is also taken as 4 (same
as in \cite{ygq03}) for each object site in our approach. This
entire configuration is to make an equitable comparison as much as
possible. Now the only difference between our Bayesian network
predictor and the Markov Chain method is that we utilize the whole
spatial and temporal information in a transportation network to
forecast while the latter only uses the temporal information of
the object site.

We take road link $Gd$ as an example to show our approach. $Gd$
represents the vehicle flow from upstream link $F$ to downstream
link $G$. All the available traffic flows which may be informative
to forecast $Gd$ in the transportation network includes $\{Ba, Bb,
Bc, Ce, Cf,  ...  , Ka, Kb, Kc, Kd\}$. Considering the time
factor,  to forecasting the traffic flow $Gd(t)$ (effect node), we
need judge the above sites with different time indices, such as
$\{Ba(t-1), Ba(t-2), ... ,Ba(t-d)\}, \{Bb(t-1), Bb(t-2), ... ,
Bb(t-d)\}$, etc. In this paper, $d$ is taken as 100 empirically.
Four most correlated traffic flows which are selected with the
correlation variable ranking criterion and the best-first strategy
for five different object traffic flows  and the corresponding
correlation coefficient values are listed in Table \ref{corcoef}.
\setlength{\tabcolsep}{4pt}
\begin{table}
\centering
 \caption{Four most correlated traffic flows for five object traffic flows}
\label{corcoef}
\begin{tabular}{l||llll}
\hline
Object traffic flows&  \multicolumn{4}{l}{ Strongly correlated  traffic flows (cause nodes)}\\
 \hline
$Ch(t)$ &$Bc(t-1)$&$Hl(t-1)$&$Ch(t-1)$&$Fe(t-3)$\\
        &0.971&0.968&0.967&0.966\\
$Dd(t)$ &$Dd(t-1)$&$Ch(t-1)$&$Bc(t-1)$&$Hl(t-2)$\\
        &0.963&0.961&0.959&0.959\\
$Fe(t)$ &$Fe(t-1)$&$Ba(t-1)$&$Fe(t-2)$&$Fe(t-96)$\\
        &0.983&0.978&0.964&0.961\\
$Gd(t)$ &$Gd(t-1)$&$Fh(t-1)$&$Hl(t-1)$&$Fe(t-1)$\\
        &0.967&0.962&0.962&0.957\\
$Ka(t)$ &$Hi(t-1)$&$Cf(t-1)$&$Ka(t-1)$&$Bb(t-2)$\\
        &0.967&0.967&0.967&0.966\\
\hline
\end{tabular}
\end{table}
\setlength{\tabcolsep}{1.4pt}
\setlength{\tabcolsep}{4pt}
\begin{table}
\centering
 \caption{A performance comparison of three methods for short-term
traffic flow forecasting of five different road links}
\label{result}
\begin{tabular}{l||lllll}
\hline
Methods& Ch & Dd & Fe & Gd & Ka \\
 \hline
Random Walk         &79.85 &70.99 &157.60 &177.57 &99.20 \\
Markov Chain        &68.51 &66.15 &122.65 &151.31 & 80.46\\
Spatio-Temporal Bayesian Network&{\it 65.95} &{\it 57.46} &{\it 115.07} &{\it 141.37} &{\it 73.02}\\
\hline
\end{tabular}
\end{table}
\setlength{\tabcolsep}{1.4pt}

With the selected cause nodes (input traffic flows), we can
approximate the joint probability distribution between the input
and output  with GMM, then derive the optimum prediction
formulation for road link $Gd$. In addition, we also conducted
experiments on four other traffic flows. Table \ref{result} gives
the forecasting errors denoted by Root Mean Square Error (RMSE) of
all the five road links through Random Walk method,  Markov Chain
method and our predictor. In the same column of Table
\ref{result}, the smaller RMSE corresponds to the better
forecasting accuracy. From the experimental results, we can find
the significant improvements of forecasting capability brought by
the spatio-temporal Bayesian network predictor which integrates
both spatial and temporal information for forecasting.
\section{Conclusions and Future Work}
In this paper, we successfully combine the whole spatial with
temporal information available in a transportation network to
carry out short-term traffic flow forecasting. Experiments show
that distant road links in a transportation network can have high
correlation coefficients, and this relevance can be employed for
traffic flow forecasting. This knowledge would greatly broaden
people's traditional knowledge about transportation networks and
the transportation forecasting research. Many existing methods can
be illuminated and further developed on the scale of a
transportation network. In the future, how to extend the presented
spatio-temporal Bayesian network predictor to forecast traffic
flows in case of incomplete data would be a valuable direction.
\section*{Acknowledgements}
The authors are grateful to the anonymous reviewers for giving
valuable remarks. This work was supported by Project 60475001 of
the National Natural Science Foundation of China.

\end{document}